\documentclass[journal,twoside,web]{ieeecolor}
\usepackage{generic}
\usepackage{cite}
\usepackage{amsmath,amssymb,amsfonts}
\usepackage{chngcntr}
\counterwithout{figure}{section}

\usepackage{algorithmic}
\usepackage{graphicx}
\usepackage{algorithm,algorithmic}
\usepackage{hyperref}
\hypersetup{hidelinks=true}
\usepackage{textcomp}
\usepackage{xcolor}
\usepackage[caption=false,font=footnotesize,labelfont=sf,textfont=sf]{subfig}
\usepackage{soul}
\usepackage{flushend}
\def\BibTeX{{\rm B\kern-.05em{\sc i\kern-.025em b}\kern-.08em
    T\kern-.1667em\lower.7ex\hbox{E}\kern-.125emX}}
\begin{document}

\title{Explanation-Aware Learning for Enhanced Interpretability in Biomedical Imaging}

\author{Zubair Faruqui, Rahul Dubey \\
\textit{Department of Computer Science, Missouri State University} \\
% \textit{}\\
Zubair10@MissouriState.edu, RahulDubey@MissouriState.edu
}

\maketitle

\begin{abstract}

Deep neural networks for medical image diagnosis often achieve high predictive accuracy while relying on spurious or clinically irrelevant visual cues, limiting their trustworthiness in practice. Post-hoc explanation methods are widely used to visualize model decisions in the form of saliency maps; however, these explanations do not influence how models learn during training, allowing non-causal or confounding features to persist. This motivates the incorporation of explanation supervision directly into the training objective to guide model attention toward clinically meaningful regions and promote clinically grounded decision-making. 
% The endeavor remains relatively underexplored, and existing approaches typically rely on a single explanation loss formulation. 
This paper presents a systematic approach to integrate explanation loss into model training and analyzes how different explanation loss designs and supervision strengths influence both predictive performance and spatial faithfulness of explanations. To quantitatively assess interpretability, two complementary explanation performance metrics—annotation coverage and saliency precision, are introduced, enabling rigorous evaluation beyond qualitative visualization. Our experimental results reveal a clear trade-off between explanation quality and explanation loss coefficients. Furthermore, quantitative statistical analysis yields consistently improved explanation alignment while maintaining comparable accuracy. Experiments were conducted on annotated chest X-ray datasets; but, the proposed framework is applicable to a broad range of annotated biomedical imaging modalities. Overall, these findings demonstrate that explanation supervision is not a monolithic design choice and provide practical guidance for incorporating explanation loss into training objectives under noisy clinical annotations.

\end{abstract}


\section{Introduction}

Deep learning models have achieved strong performance in medical image diagnosis across modalities such as chest X-rays, computed tomography (CT), and magnetic resonance imaging (MRI) \cite{rajpurkar2017chexnet, litjens2017survey}. Despite these advances, clinical adoption remains limited due to a lack of transparency and reliance on spurious visual cues \cite{zech2018variable, badgeley2019deep}. In safety-critical healthcare settings, predictive accuracy alone is insufficient, and models must provide explanations aligned with clinical reasoning to enable trust and responsible deployment \cite{kelly2019key, tonekaboni2019clinicians}. To interpret deep model's behavior, explainable artificial intelligence (XAI) techniques have emerged as a key mechanism for improving model decision transparency \cite{guidotti2018survey,tjoa2020medical,faruqui2025explainabilitycnnbasedclassification,doshi2017rigorous,samek2019xai}. Among XAI approaches, gradient-based saliency methods such as Gradient-weighted Class Activation Mapping (Grad-CAM) \cite{selvaraju2017gradcam} is widely used to generate visual explanations by highlighting image regions that contribute most strongly to model predictions. Grad-CAM has been applied extensively in radiology to visualize disease-relevant regions and to assist clinicians in interpreting model outputs \cite{saporta2022benchmarking}. However, post-hoc explanations do not influence how models learn during training, and they may still base their predictions on confounding or non-causal features while making accurate predictions.

Recent studies have started exploring the incorporation of explanation supervision directly into model training by approximating the quality of explanation using expert-provided annotations \cite{sun2020explanation, uddin2025expert}. These approaches demonstrate that guiding model attention during training can improve interpretability and maintain competitive predictive performance\cite{rieger2020interpretationsusefulpenalizingexplanations, ross2017right}, particularly in medical imaging tasks where expert knowledge is available. However, existing explanation-guided training approaches lack in interpreting the quality of explanation in a quantitative manner as a result of explanation loss incorporation. 
As a result, the impact of different explanation loss formulations and their effect on the balance between interpretability and predictive performance remains insufficiently understood. Clarifying this relationship is critical for developing explainable models that are both reliable and clinically meaningful. 

This paper presents an explanation-guided training framework and systematically analyzes how different explanation loss formulations and their weighting influence explanation quality and model performance. We evaluated the performance of a proposed explanation-aware training method for medical image classification using the VinDr-CXR chest X-ray dataset~\cite{Nguyen2022VinDrCXR} through a comprehensive experimental study that span over seven disease categories. Our results show that explanation supervision can substantially improve the spatial faithfulness of saliency explanations without significantly degrading predictive performance. In particular, different explanation loss formulations, such as logit-based and probability-based, lead to markedly different explanation behaviors. We further observe that the strength of explanation supervision governs a clear trade-off between different explanation performance metrics, enabling explanation behavior to be tuned according to clinical requirements and annotation granularity.

To rigorously assess interpretability and explanation quality, we introduce two quantitative localization metrics: 1) annotation coverage and 2) saliency precision, complementing qualitative visualization. These metrics help in investigating whether improvements arise from broader attention, sharper focus, or genuine alignment with disease-relevant regions. Through extensive empirical analysis, we show that explanation supervision exhibits heterogeneous effects depending on loss formulation and explanation loss coefficient. The logit-based losses consistently yield more faithful and stable saliency maps than probability-based alternatives. Moreover, results show that the explanation loss coefficient governs a trade-off between different aspects of explanation quality.

% enabling explanation behavior to be tuned without significantly degrading classification accuracy.

The main contributions of this work are summarized as follows: (1) We introduce a unified explanation-aware training framework that jointly optimizes classification performance and explanation quality by augmenting binary cross-entropy loss with multiple explanation-driven objectives. (2) We provide a principled characterization of the trade-off between explanation loss weighting and explanation quality, offering guidance for balancing predictive accuracy and interpretability during training.
(3) We establish, through statistical quantitative evaluation, that the proposed framework yields robust and consistent explanations across multiple disease categories.
Our findings provide practical guidance for designing explanation-aware training objectives in medical imaging, particularly under noisy spatial supervision. By revealing how loss formulation and coefficients shape both prediction and explanation behavior, this work advances the understanding of how explainability can be meaningfully integrated into model training, moving beyond post-hoc visualization toward more trustworthy and clinically aligned AI systems.

\section{Background and Related Work}
\label{sec:related_work}

% \subsection{Explainable AI in Medical Imaging}
% \label{sec:xai_medical}

XAI has become an essential component of medical AI systems, aiming to provide transparency into model decision-making processes \cite{loh2022xaihealthcare}. Among post-hoc explanation techniques, gradient-based saliency methods are widely used due to their architectural flexibility and ease of application. The most popular among these, Grad-CAM produces class-discriminative heatmaps by leveraging gradients flowing into convolutional feature maps, which has been widely applied in radiology to visualize disease-relevant regions and assist clinicians in interpreting model outputs, including visualization of pathological areas \cite{panwar2020deep}, heatmap localization in chest X-ray classification tasks \cite{umair2021detection}, and multi-modal anomaly localization in brain MRI and chest X-rays \cite{erukude2025explainable}. Additionally, Grad-CAM has been evaluated as part of saliency benchmarking studies in medical imaging, further highlighting its prominence in the field \cite{saporta2022benchmarking}. Beyond gradient-based approaches, other explanation paradigms such as perturbation-based methods (e.g., LIME and SHAP) and attribution-based techniques (e.g., Layer-wise Relevance Propagation, LRP) have also been explored in medical imaging, each offering distinct trade-offs in terms of fidelity, stability, and computational cost \cite{ribeiro2016lime,lundberg2017shap,lapuschkin2019lrp}.

% \subsection{Limitations of Post-hoc Explanations}
% \label{sec:posthoc_limits}

However, prior studies have shown that saliency-based explanations can be sensitive to model parameters, data artifacts, and architectural choices, underscoring the need for careful validation before clinical use \cite{kindermans2017reliability,adebayo2018sanity}. A fundamental limitation of post-hoc explanation methods is that they do not influence the training process. A model may achieve high predictive accuracy while relying on spurious correlations or dataset biases. 
% yet still generate visually plausible explanations that appear convincing to non-expert observers, even when the highlighted regions are not truly relevant to the model’s underlying decision process. (\textcolor{red}{how is this possible == generate visually plausible explanations but explanation is not good?}). 
This disconnect has motivated increasing interest in methods that encourage models to be \textit{right for the right reasons} by explicitly constraining explanatory behavior during training \cite{ross2017right}. Several studies demonstrate that saliency maps may fail to reliably indicate causal evidence and explanation can still remain unreliable even when model parameters are re-tuned, raising concerns about their faithfulness as indicators of model reasoning \cite{adebayo2018sanity,kindermans2017reliability}. These observations motivate explanation-aware training strategies that shape model attention during optimization rather than relying exclusively on post-hoc interpretation.

% \subsection{Explanation-Guided Training and Attention Supervision}
% \label{sec:explanation_guided}

To address these limitations, recent work has explored incorporating explanation signals directly into model training objectives. Ross \cite{ross2017right} proposed constraining model gradients to align explanations with human-specified rationales, establishing a foundational framework for explanation regularization. Subsequent studies extended this idea by using explanation signals to guide training and improve generalization in few-shot and cross-domain learning settings~\cite{sun2020explanation}. 
Šefčík \cite{sefcik2021lrp} employed LRP to analyze and encourage model relevance with respect to brain tumor regions in MRI under full-data training regimes. Similarly, Caragliano\cite{caragliano2021doctor} proposed the Doctor-in-the-Loop framework, which integrates expert-provided segmentation masks to guide model learning for non-small cell lung cancer analysis in CT images.

In the medical imaging domain, expert-guided attention supervision has been applied using region-of-interest annotations to encourage models to focus on diagnostically meaningful areas. A recent representative example integrates Grad-CAM alignment with expert annotations within a few-shot learning framework, demonstrating improvements in both interpretability and classification accuracy \cite{uddin2025expert}. While effective, such approaches typically adopt a single explanation loss formulation and a fixed supervision strength, leaving the broader design space of explanation-aware objectives largely unexplored. As a result, little is known about how different explanation-aware loss designs influence model behavior, how explanation strength should be weighted during training, or how these choices affect the trade-off between predictive performance and spatial faithfulness of explanations.
% \subsection{Challenges in Saliency Evaluation with Coarse Annotations}
% \label{sec:coarse_annotations}

Another issue is that often annotations provided by experts are coarse (large bounding boxes) rather than pixel-level segmentation. These bounding boxes may enclose large regions of normal anatomical area, while the actual pathology occupies only a small fraction of the annotated area. 
% Consequently, strict overlap-based evaluation metrics may underestimate explanation quality, particularly for small or diffuse abnormalities.
Recent large-scale benchmarking studies emphasize that many saliency methods perform inconsistently when evaluated quantitatively and caution against relying solely on qualitative visualization for clinical validation \cite{saporta2022evaluation}. Related work in weakly supervised disease localization similarly highlights that annotation granularity and evaluation protocol choices substantially influence conclusions about localization and interpretability performance \cite{li2018thoracic}.

% \subsection{Positioning of the Present Work}
% \label{sec:positioning}

Despite growing interest in explanation-guided training, two important gaps remain. First, there is a limited understanding of how different explanation-aware loss formulations influence saliency behavior and model learning. Second, the relationship between explanation supervision strength and the resulting coverage--precision trade-off under coarse clinical annotations has not been quantitatively studied.
In this work, we address these gaps by conducting a comprehensive investigation of explanation-guided training objectives for medical image classification problem. We evaluated seven explanation-aware loss formulations across four different explanation strengths and seven disease categories, and introduce quantitative localization metrics tailored to coarse bounding box annotations. 
% Table~\ref{tab:prior_work} summarizes how the proposed approach differs from and extends prior explanation-guided training methods.

% \begin{table}[t]
% \caption{Comparison of prior explanation-guided training approaches with the proposed method.}
% \label{tab:prior_work}
% \centering
% \begin{tabular}{lcc}
% \hline
% \textbf{Aspect} & \textbf{Prior Work} & \textbf{This Work} \\
% \hline
% Grad-CAM used during training & Yes & Yes \\
% Explanation loss formulations & Single & Multiple (8) \\
% Logit vs probability supervision & No & Yes \\
% Explanation coefficient ablation & Limited & Systematic (4) \\
% Diseases evaluated & Few & Seven \\
% Quantitative saliency metrics & Limited & Coverage \& Precision \\
% Coarse box robustness analysis & No & Yes \\
% \hline
% \end{tabular}
% \end{table}

\section{Methodology}
\subsection{Dataset and CNN Model}
In this work, experiments are conducted on the VinDr-CXR dataset\cite{Nguyen2022VinDrCXR}, a large-scale, publicly available chest X-ray radiographs with radiologist-provided annotations in the form of bounding boxes that indicate approximate regions of abnormality. These annotations are inherently coarse and may include substantial regions of normal anatomy or omit portions of the true pathology. Despite these limitations, the objective is to leverage this weak spatial supervision during model training to guide its attention toward clinically relevant regions while maintaining strong classification performance.

% \textcolor{red}{details of the dataset, its multi class and multi label dataset that has 14 different disease classes. Also mention how many samples in total, that belong to normal and abnormal class.}
This dataset is a multi-label dataset comprising 14 thoracic disease categories, along with a \emph{No Finding} label indicating normal cases. It contains over 18{,}000 chest X-ray images, of which a substantial subset exhibits one or more annotated abnormalities, while the remaining images correspond to normal findings. Each image may be associated with multiple disease labels, reflecting the presence of co-occurring pathologies.

From this dataset, we construct seven binary classification tasks corresponding to the most prevalent disease categories: \emph{Cardiomegaly}, \emph{Pleural effusion}, \emph{Lung opacity}, \emph{Aortic enlargement}, \emph{Atelectasis}, \emph{Consolidation}, and \emph{Pleural thickening}. 
% (\textcolor{red}{1, 2.. 7}). 
% For each task, the presence of a specific disease is treated as the positive class, while images labeled as \emph{No Finding} are used as negative samples. 
To mitigate class imbalance, the negative class is subsampled to match the number of disease-positive samples for each task. Across diseases, the number of positive samples ranges from approximately 1{,}000 to 3{,}000 samples per task.

To accurately predict these diseases, we employ a pre-trained DenseNet-121 classification model\cite{Huang2017DenseNet}. The pretrained network is adapted for binary classification by replacing the final classifying layer with a two-neuron linear layer preceded by dropout, which was hand-tuned to reduce overfitting.
% (\textcolor{red}{why dropout, we hand tuned the model architecture to reduce overfitting)}, \textcolor{red}{find the default dropout value/rate from DenseNet-121: 0.2 within the conv layers only}. 
% All network layers are unfrozen during training to allow end-to-end optimization and to ensure that gradients can propagate through the final convolutional layer, which is required for explanation heatmap computation. Activations from the last convolutional block are captured using a forward hook and are later used for gradient-based explanation loss computation and post-hoc interpretability analysis. 
All network layers are unfrozen during training to enable end-to-end optimization and to ensure that gradients can propagate through the final convolutional layer, which is required for explanation heatmap computation. The model is initialized with pretrained weights, as prior work has shown that pretrained initialization consistently outperforms random weight initialization even when models are fully fine-tuned end-to-end \cite{rajaraman2024transfercxr}.

% \subsection{Implementation and Training Details}

The model was trained for 60 epochs using the Adam optimizer with a learning rate of $2 \times 10^{-4}$ and weight decay of $1 \times 10^{-4}$. A dropout rate of 0.3 is applied to the classification head to improve generalization. Due to the high input resolution and the additional memory overhead introduced by Grad-CAM-based explanation supervision, a batch size of 12 is used during training and evaluation. All experiments use identical batch sizes to ensure fair comparison across loss formulations and supervision strengths. 
% The input images are resized to $1024 \times 1024$ using letterboxing to preserve aspect ratio. 
To systematically study the effect of explanation-aware training, we train models using eight different loss formulations and four explanation loss coefficients for each disease category, resulting in a total of 224 trained models.

% Let $f_{\theta}(x)$ denote a convolutional neural network parameterized by $\theta$ that produces logits
% % \[
% % \mathbf{z} = [z_0, z_1],
% % \]

% corresponding to the negative and positive classes, respectively. 
\subsection{Training Objective}

The model training objective has three components: a standard binary classification loss, an explanation-aware penalty, and explanation loss coefficient ($\alpha$) term, as shown in Eq.~\eqref{eq:total_loss}. The total training loss is defined as
\begin{equation}
\mathcal{L}_{\text{total}} = \mathcal{L}_{\text{bce}} + \alpha \, \mathcal{L}_{\text{exp}}
\label{eq:total_loss}
\end{equation}
where $\mathcal{L}_{\text{bce}}$ denotes the binary cross-entropy loss, $\mathcal{L}_{\text{exp}}$ represents the explanation-aware loss derived from Grad-CAM using different loss formulations, and $\alpha$ is an explanation loss coefficient that controls the strength of explanation supervision during training.

\subsubsection{Binary Cross-Entropy Loss}
let $y \in \{0,1\}$ denote the ground-truth class label and $\hat{p} \in [0,1]$ denote the predicted probability for the disease-positive class. The binary cross-entropy loss is defined as in Eq~\ref{eq:bce_loss}.
\begin{equation}
\mathcal{L}_{\text{bce}} = - \left[ y \log(\hat{p}) + (1 - y)\log(1 - \hat{p}) \right].
\label{eq:bce_loss}
\end{equation}

\subsubsection{Explanation Based Loss}
While the BCE loss optimizes classification performance, explanation-aware training requires a scalar output whose gradients can be used to generate class-discriminative attribution maps. In the Grad-CAM framework, this scalar score, $(s)$, determines how gradients are propagated to the final convolutional feature maps and thus directly influences the resulting heatmaps. To study how different choices of this scalar affect explanation quality, we systematically explore multiple explanation score formulations. We also consider a standard classification baseline trained using only a BCE loss, without any explanation score. This \emph{Pure BCE} model serves as an orthodox reference point for evaluating the impact of explanation-guided training. 

We adopt Grad-CAM as the explanation mechanism for explanation-aware training due to its architectural compatibility with convolutional neural networks and its ability to produce class-discriminative spatial attributions without requiring additional model components. Unlike perturbation-based or surrogate explanation methods, Grad-CAM provides gradient-based localization that can be efficiently integrated into the training objective and differentiated end-to-end. 

In the following logit-based and probability-based formulations, the class value difference explicitly increases evidence for the positive class while suppressing evidence for the negative class. This joint effect encourages separation between classes, yielding explanations that are inherently contrastive and emphasize features that discriminate between positive and negative predictions.
% (\textcolor{red}{Need explanation:  Z1- Z0 push the model more towards positive class and away from negative class. This makes the explanations more contrastive and less discriminative.})
These scores can be computed in following ways.

\paragraph{Logit-based scores.}

Let $z_0$ and $z_1$ denote the logits corresponding to the negative and positive classes, respectively. We define the following logit-based explanation scores:

\[
s_{\text{logit}}^{\text{alg}} = z_1 - z_0, \quad
s_{\text{logit}}^{\text{abs}} = \lvert z_1 - z_0 \rvert, \quad
s_{\text{logit}}^{\text{sqr}} = (z_1 - z_0)^2
\]

where the three scores correspond to the logit algebraic, logit absolute, and logit squared differences, respectively.

For class discriminative explanation analysis, we also considered the score for the positive class as well:

\[
s_{\text{logit}} = z_1
\]

which uses only the disease-class logit without contrast to the negative class. This formulation serves to assess the importance of class-contrastive signals for explanation-guided training.

\paragraph{Probability-based scores.}

Let $p_0$ and $p_1$ denote the softmax probabilities for the negative and positive classes, respectively. The corresponding probability-based explanation scores are defined as:

\[
s_{\text{prob}}^{\text{alg}} = p_1 - p_0, \quad
s_{\text{prob}}^{\text{abs}} = \lvert p_1 - p_0 \rvert, \quad
s_{\text{prob}}^{\text{sqr}} = (p_1 - p_0)^2
\]

where the three scores correspond to the probability algebraic, probability absolute, and probability squared differences, respectively.

The explanation loss is applied only to disease-positive samples with available bounding box annotations, ensuring that explanation supervision is grounded in expert-provided spatial information.
For each disease-positive sample, we compute Grad-CAM heatmaps using activations from the final convolutional layer. Let
\(
\mathbf{A} \in \mathbb{R}^{c \times h \times w}
\)
denote the feature maps of this layer, where $c$, $h$, and $w$ denote the channel, height, and width dimensions, respectively. Let
\(
s
\)
be a scalar score derived from the model output, the Grad-CAM weight ($\alpha_c$) is computed using Eq~\ref{eq:grad_cam_weight}.
\begin{equation}{\label{eq:grad_cam_weight}}
\alpha_c
=
\frac{1}{h w}
\sum_{i=1}^{h}
\sum_{j=1}^{w}
\frac{\partial s}{\partial A^{c}_{ij}},
\end{equation}
where \( A^{c}_{ij} \) denotes the activation at spatial location \((i,j)\) of channel \(c\).
The resulting Grad-CAM heatmap (H) is then obtained as given in Eq~\ref{eq:H}.
\begin{equation}{\label{eq:H}}
\mathbf{H}
=
\mathrm{ReLU}
\left(
\sum_{c}
\alpha_c \mathbf{A}^{c}
\right).
\end{equation}

% For each disease-positive sample, we compute Grad-CAM heatmaps using activations from the final convolutional layer. Let
% \[
% \mathbf{A} \in \mathbb{R}^{C \times h \times w}
% \]
% denote the feature maps of this layer, where $C$ is the number of channels and $h \times w$ is the spatial resolution. Let $s$ denote a scalar score derived from the model output, as defined in Section~3.4.

% The Grad-CAM weights are computed as
% \[
% \alpha_c = \frac{1}{hw} \sum_{i=1}^{h} \sum_{j=1}^{w}
% \frac{\partial s}{\partial A^{c}_{ij}},
% \]
% where $A^{c}_{ij}$ denotes the activation at spatial location $(i,j)$ in channel $c$.

% The resulting Grad-CAM heatmap is given by
% \[
% \mathbf{H} = \mathrm{ReLU}\left( \sum_{c} \alpha_c \mathbf{A}^{c} \right).
% \]

% The heatmap is subsequently min--max normalized to the range $[0,1]$ to allow consistent comparison across samples.

% \subsection{Explanation Score Design}
% \label{sec:explanation_score}

\begin{figure}[t]
    \centering
    \subfloat[Expert-provided bounding box annotations]{
        \includegraphics[width=0.45\columnwidth]{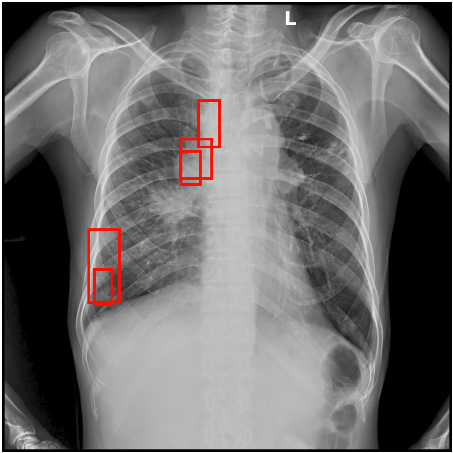}
        \label{fig:multi_box}
    }\hfill
    \subfloat[Union mask used for explanation supervision]{
        \includegraphics[width=0.45\columnwidth]{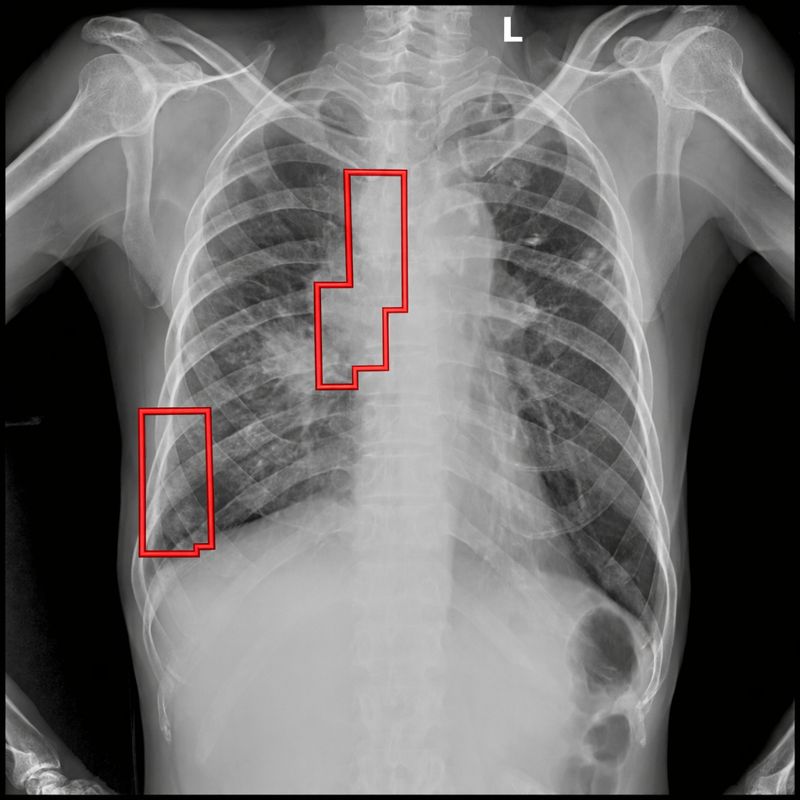}
        \label{fig:union_mask}
    }
    \caption{Construction of the annotation mask $M$ from expert-provided bounding boxes.
    (a) Multiple coarse bounding box annotations provided by radiologists for a disease-positive chest X-ray. Individual boxes may cover large regions containing both pathological and normal tissue.
    (b) Binary union mask $M$ (in red) formed by merging all bounding boxes into a single spatial supervision signal aligned with the Grad-CAM resolution. This union-based representation enables consistent explanation supervision in the presence of multiple and overlapping annotations.}
    \label{fig:box_mask_union}
\end{figure}

% \subsection{Spatial Non-Overlap Loss}
% \label{sec:cam_loss}
To incorporate expert spatial annotations, the provided bounding boxes are converted into a binary mask 
$M \in \{0,1\}^{h \times w}$ aligned with the spatial resolution of the saliency heatmap. 
Bounding boxes defined in the input image space are mapped to the saliency resolution using the same spatial scaling applied to the feature maps. 
When multiple bounding boxes are present for a single image, the mask $M$ is constructed as the union of all annotated boxes, such that 
$M_{ij} = 1$ if pixel $(i,j)$ lies inside at least one bounding box, and $M_{ij} = 0$ otherwise. This union-based representation enables a unified treatment of multiple expert annotations. A schematic illustration of this process is shown in Fig.~\ref{fig:box_mask_union}.

Rather than enforcing strict overlap between the saliency heatmap and the full extent of the bounding box, we focus on the concentration of salient regions. This design choice is motivated by the coarse nature of radiologist annotations, which often include substantial regions of normal tissue while the actual pathology occupies only a small fraction of the annotated area. Enforcing full overlap would therefore bias the model toward diffuse attention rather than meaningful localization.

Let $H \in \mathbb{R}^{h \times w}$ denote the raw saliency heatmap. 
We normalize it to the range $\hat{H} \in [0,1]$.
% as
% \begin{equation}
% \hat{H}(i,j) = \frac{H(i,j) - \min(H)}{\max(H) - \min(H) + \epsilon},
% \label{eq:gradcam_norm}
% \end{equation}
% where $\epsilon$ is a small constant for numerical stability.
Min--max normalization is applied to ensure comparability of saliency magnitudes across samples and training iterations, preventing explanation loss values from being dominated by scale variations in raw Grad-CAM activations. This normalization also stabilizes optimization when the explanation loss is combined with the classification objective.

To emphasize on informative regions, suppress low-magnitude noise in the explanation/heatmap, and for quantitative analysis of results, we retain only the top $k\%$ of the normalized saliency heatmap when computing the explanation loss. This quantile-based thresholding focuses supervision on the most salient regions identified by the model, rather than enforcing alignment over diffuse background activations. $k$ is set to $50$, retaining the top half of saliency activations when computing the explanation loss. We also tried other values of $k$, but set 50 as this gave more promising results. This choice reflects a balance between suppressing low-magnitude noise and allowing sufficient spatial flexibility given the coarse nature of bounding box annotations.
We define a binary thresholded saliency map ($\hat{H}^+ \in \{0,1\}^{h \times w}$) 
by retaining the top $k\%$ most salient pixels of $\hat{H}$:
\begin{equation}
\hat{H}^+(i,j) =
\begin{cases}
1, & \text{if } \hat{H}(i,j) \ge \theta_k, \\
0, & \text{otherwise},
\end{cases}
\label{eq:gradcam_threshold}
\end{equation}
where $\theta_k$ denotes the $k$-th quantile of normalized map ($\hat{H}$).
 % \textcolor{red}{Can you show $H^$ and $H^{+}$ and M.. To Do in code}. 
Finally, the explanation loss is computed using Eq~\ref{eq:explanation_loss}.
\begin{equation}
\mathcal{L}_{\text{exp}} =
1 -
\frac{\sum \left( \hat{\mathbf{H}}^{+} \odot \mathbf{M} \right)}
{\sum \hat{\mathbf{H}}^{+}}
\label{eq:explanation_loss}
\end{equation}

where $\odot$ denotes element-wise multiplication.
This loss penalizes saliency mass that falls outside expert-annotated regions, encouraging the model to allocate its most influential activations within disease-relevant areas. This formulation is robust to noisy annotations, as it does not require the entire bounding box to correspond to pathology and allows partial coverage.

\subsubsection{Explanation Loss Coefficient ($\alpha$)}
\label{sec:alpha_interpretation}

% The coefficient parameter $\alpha$ controls the relative influence of the explanation-aware loss with respect to the classification objective during training. Because expert-provided bounding box annotations are inherently coarse and may not precisely delineate pathological regions, the appropriate strength of explanation supervision is not known a priori. An overly weak explanation loss may have limited impact on model attention, while an excessively strong constraint may bias the model toward over-localized or overly restrictive explanations.

% To account for this uncertainty, 
We treat $\alpha$ as a tunable hyperparameter that governs how strongly explanation loss-based supervision shapes the learned representations. By adjusting $\alpha$, the training process can balance learning discriminative features for classification against aligning model attention with expert-annotated regions. In our experiments, we evaluate four values of $\alpha \in \{0.25, 0.50, 0.75, 1.00\}$ to systematically study how varying the strength of explanation supervision influences model behavior. The impact of this coefficient on predictive performance and explanation quality is analyzed in the Results section.

% \subsection{Implementation Notes}
% \label{sec:alpha_interpretation}

% Grad-CAM gradients are computed using automatic differentiation with retained computation graphs to allow backpropagation through the explanation loss. The CAM loss is evaluated in full precision to ensure numerical stability, and its gradients are propagated jointly with the classification loss during training.

\subsection{Explanation Evaluation Metrics}

% \st{Evaluating visual explanations in medical imaging is challenging due to coarse expert annotations and variability in saliency map structure. Moreover, different explanation behaviors may be desirable depending on the clinical context, such as broad localization of abnormal regions versus precise focus on the most informative subregions.}
% Evaluation of visual explanations in medical imaging is complicated by coarse annotations and diverse saliency patterns. Moreover, the desired explanation behavior may vary by clinical context, ranging from broad localization to precise regional focus.
% To capture these complementary aspects of explanation quality, 

We considered three quantitative evaluation metrics that assess distinct properties of saliency explanations. Specifically, we measure (a) how much of the most salient attribution mass lies within expert-provided regions (top), (b) how the overall saliency distribution aligns with annotated regions (all),  (c) whether salient regions intersect clinically annotated areas (annotation). Together, these metrics enable a nuanced analysis of explanation behavior beyond qualitative visualization, allowing us to distinguish between diffuse attention, focused localization, and meaningful alignment with disease-relevant regions.

\subsubsection{Top Saliency Precision}
\label{sec:top_saliency_precision}

This metric measures the fraction of the most salient heatmap pixels that lie within expert-annotated disease regions. This metric focuses on whether the model’s strongest explanatory signals are spatially aligned with clinically relevant areas. We use the same thresholded saliency map $\hat{H}^+$, defined as a binary saliency mask obtained by thresholding the normalized saliency heatmap $\hat{H}$ at the $k$-th quantile. .

\begin{equation}
\text{Precision}_{\text{top}} =
\frac{
\sum_{i,j} \hat{H}^+(i,j)\, M(i,j)
}{
\sum_{i,j} \hat{H}^+(i,j)
}.
\label{eq:top_saliency_precision_eq}
\end{equation}

% This metric evaluates whether the most salient explanatory regions identified by the model are concentrated within disease-relevant areas, while remaining robust to coarse bounding box annotations.

\subsubsection{All Saliency Precision }
\label{sec:all_saliency_precision}

% To complement top saliency precision with a threshold-free measure, we compute \emph{All Saliency Precision}. 
Unlike top saliency precision, this metric operates directly on the continuous normalized saliency map $\hat{H}$ without binarization, capturing the distribution of the full attribution mass.
This metric quantifies the fraction of the total saliency mass that lies within expert-annotated regions:
\begin{equation}
\text{Precision}_{\text{all}} =
\frac{
\sum_{i,j} \hat{H}(i,j)\, M(i,j)
}{
\sum_{i,j} \hat{H}(i,j)
}.
\label{eq:all_saliency_precision}
\end{equation}

All Saliency Precision captures the overall alignment between the explanation saliency distribution and annotated disease regions, penalizing diffuse explanations even when a subset of highly salient pixels overlaps with pathology. Together, top and all saliency precisions provide complementary views of explanation quality, measuring both peak saliency localization and global saliency concentration.

\subsubsection{Annotation Coverage Metric}
\label{sec:box_coverage_metric}
Different from saliency-based metrics, annotation coverage evaluates whether a saliency explanation highlights any clinically relevant portion of an annotated region, rather than requiring dense or precise overlap. In simple terms, this metric measures whether a Grad-CAM heatmap acknowledges the presence of a clinically annotated region.
For each annotated bounding box $\mathcal{B}_k$, as shown in Fig~\ref{fig:box_mask_union}, we compute the fraction of pixels within the box that are marked as salient. A bounding box is considered \emph{covered} if
\begin{equation}
\frac{1}{|\mathcal{B}_k|}
\sum_{(i,j) \in \mathcal{B}_k}
\hat{\mathbf{H}}^{+}(i,j)
\ge \tau_{\text{cov}},
\label{eq:annotation_coverage_condition}
\end{equation}

where $|\mathcal{B}_k|$ denotes the area of the $k$-th bounding box and $\tau_{\text{cov}} = 0.01$ in our experiments. This relaxed criterion reflects clinical reality: only a small fraction of a coarse annotation may correspond to true pathology, and highlighting even a limited but relevant subregion can provide meaningful interpretability. The final annotation coverage score is reported as the fraction of covered bounding boxes over all annotated boxes.

\section{Results and Discussion}

This section presents a comprehensive empirical evaluation of the proposed explanation-aware training approach. We investigate how incorporating different explanation loss at varying loss coefficient ($\alpha$) into the training objective influences optimization behavior (training loss), predictive performance (accuracy), and the spatial faithfulness of model explanations.

Results are reported across seven disease categories and aggregated over all test samples to capture trends rather than disease-specific effects. We first analyze training dynamics to verify that integrating explanation supervision does not destabilize optimization or degrade validation accuracy. We then present qualitative comparisons of saliency heatmap explanations to visually illustrate differences between standard BCE training and explanation-guided model training.

\begin{figure*}[t]
    \centering
    \subfloat[Training loss dynamics]{
        \includegraphics[width=0.99\columnwidth]{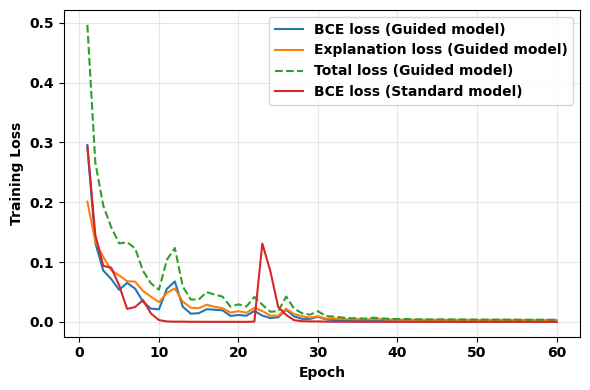}
        \label{fig:training_loss_comparison}
    }
    \hfill
    \subfloat[Validation accuracy]{
        \includegraphics[width=0.99\columnwidth]{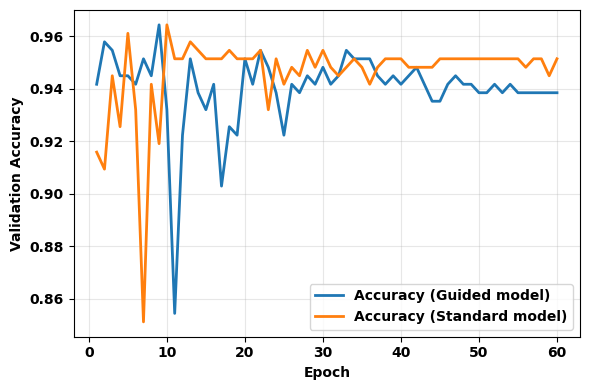}
        \label{fig:training_accuracy_comparison}
    }
    \caption{(a) Comparison of training dynamics between explanation-guided and standard BCE-trained models for the \emph{pleural effusion} disease classification task ($\alpha = 0.25$), using the \textit{Logit Difference Square} formulation.
     The guided model minimizes a weighted sum of its BCE loss and explanation penalty terms, while the standard model optimizes BCE loss alone. The close alignment between the guided model’s total loss and the standard model’s BCE loss indicates that explanation supervision does not destabilize training.
    (b) Validation accuracy comparison between the two models, converging to comparable accuracy, indicating similar predictive performance.}
    \label{fig:training_and_accuracy_comparison}
\end{figure*}

Next, we examine the effect of the explanation loss coefficients, $\alpha$, on classification performance and localization behavior, followed by an analysis of the trade-off between annotation coverage and saliency precision induced by varying supervision strength. We subsequently compare multiple explanation loss formulations to assess how design choices in explanation-aware objectives influence predictive accuracy, saliency concentration, and robustness under coarse clinical annotations. 

% Finally, we summarize the consistency of these findings across disease categories to demonstrate the robustness and general applicability of the proposed framework.

\subsection{Effect of Explanation Loss on Model Performance}
\label{sec:training_dynamics}

\subsubsection{Quantitative Comparison of training loss and accuracy}
% \label{sec:qualitative_comparison}

The model was trained using standard BCE loss and using explanation-augmented loss. Figure~\ref{fig:training_loss_comparison} illustrates the training dynamics of the explanation-guided model. The ``BCE loss (Guided Model)'' curve corresponds to the classification loss term $\mathcal{L}_{\text{bce}}$, the ``Explanation loss'' denotes the weighted explanation loss penalty $\alpha \mathcal{L}_{\text{exp}}$, and the ``total loss'' represents their sum, i.e., the full optimization objective defined in Eq.~\ref{eq:total_loss}. The additional ``BCE loss 
(Standard Model)'' curve shows the classification loss of a baseline model trained 
using BCE alone, without explanation supervision.

For the explanation-guided model, the BCE loss and the explanation loss decrease jointly over training, and their weighted sum converges smoothly. Importantly, the BCE component of the guided model follows a trajectory comparable to that of the standard model, suggesting that explanation supervision does not interfere with learning discriminative features. Instead, it acts as a regularizing signal that reshapes gradient flow while preserving stable optimization.

As shown in Fig.~\ref{fig:training_accuracy_comparison}, the explanation-guided model achieves validation accuracy comparable to the standard BCE model throughout training. Although the guided model exhibits slightly higher variance during early epochs due to the additional explanation supervision, both models converge to similar accuracy levels. A small reduction in peak accuracy (approximately 1\%) is observed for the guided model; however, this marginal loss is subsequently compensated by substantial gains in explanation quality, as demonstrated by the quantitative localization metrics and qualitative visual analyses presented in the following sections.

Overall, these results demonstrate that explanation-aware training remains well-behaved from an optimization perspective and can be integrated into standard CNN training pipelines without sacrificing convergence stability or classification accuracy.

\begin{figure}[t]
    \centering
    \subfloat[Standard training]{
        \includegraphics[width=0.45\columnwidth]{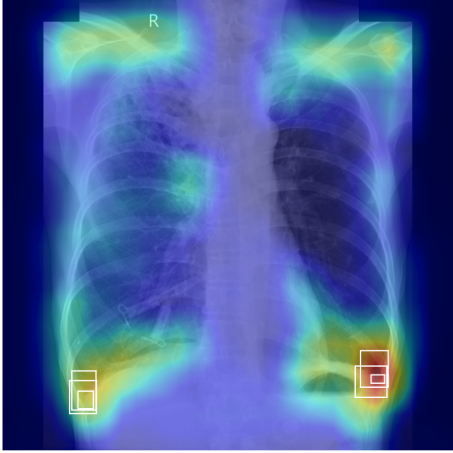}
        \label{fig:purebce_cam}
    }\hfill
    \subfloat[Explanation-guided training]{
        \includegraphics[width=0.45\columnwidth]{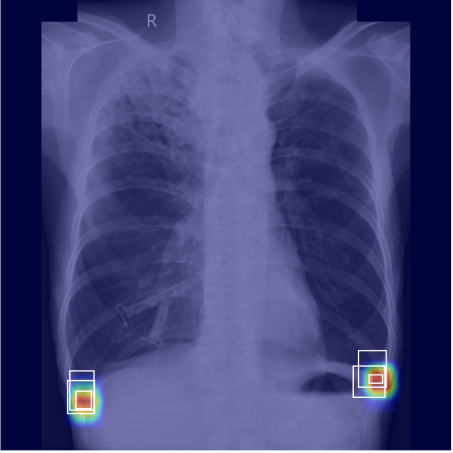}
        \label{fig:expl_guided_cam}
    }
    \caption{Qualitative comparison of Grad-CAM explanations for pleural effusion. 
    \textbf{(a)}: standard training produces diffuse saliency spanning large regions of the image.
    \textbf{(b)}: the proposed explanation-guided model yields focused localization aligned with expert-annotated disease regions (white boxes).}
    \label{fig:qualitative_comparison}
\end{figure}

\subsubsection{Qualitative Comparison of Heatmaps}
% \label{sec:qualitative_comparison}

Figure~\ref{fig:qualitative_comparison} provides a visual qualitative comparison of heatmaps generated using Grad-CAM when the CNN model was trained considering standard BCE training (a) and the proposed explanation-guided approach (b), respectively. 
When trained with BCE alone, the model produces highly diffuse Grad-CAM responses that span large portions of the scan, leading to good annotation coverage but poor spatial specificity.
In contrast, explanation-guided training encourages saliency concentration within clinically relevant regions, yielding sharper and more localized explanations that better align with expert annotations.
This qualitative behavior is consistent with the quantitative trends observed in annotation coverage and saliency precision metrics described in the next sections. We conducted extensive experiments covering all disease types, considering several different samples, and observed similar qualitative and quantitative performance.

\begin{figure*}[t]
    \centering
    \subfloat[Effect of $\alpha$ on predictive performance]{
        \includegraphics[
            width=0.99\columnwidth,
            height=5cm
        ]{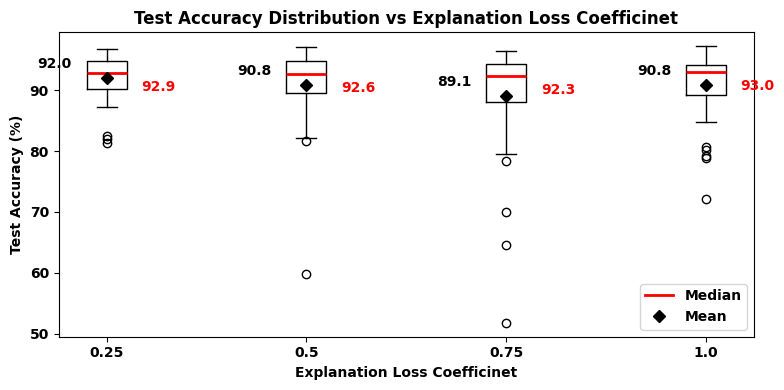}
        \label{fig:WeightAccuracyCoverage}
    }
    \hfill
    \subfloat[Effect of $\alpha$ on annotation coverage and saliency precision]{
        \includegraphics[width=0.99\columnwidth, height=5cm]{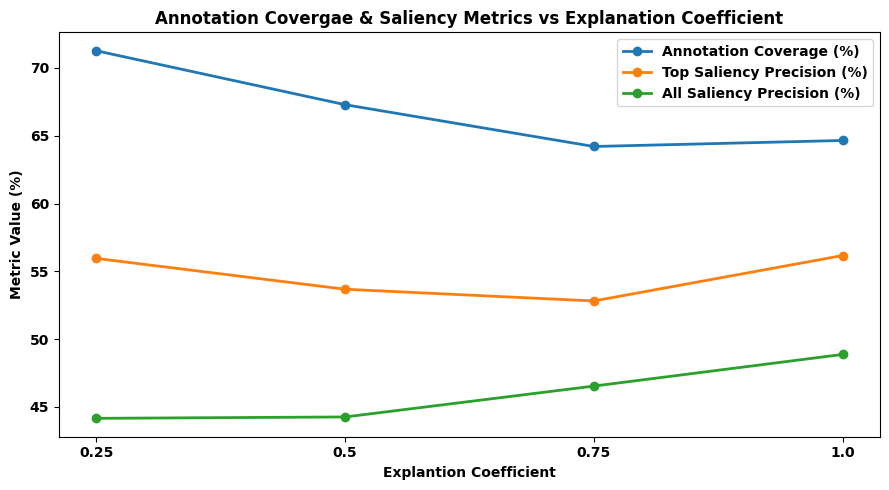}
        \label{fig:SaliencyPrecisionByWeights}
    }
    \caption{Effect of explanation loss coefficient on predictive performance (a) and explanation quality across seven disease categories (b). In fig (a), box plots summarize distributions of test accuracies obtained over different values of $\alpha$ for seven loss formulations and for seven diseases, a total of $49$ test accuracy values. As $\alpha$ increases, the mean accuracy decreases, but median remains stable. Fig (b) shows average annotation coverage, top-saliency precision, and all-saliency precision over 49 experimental settings. As $\alpha$ increases, annotation coverage and top-saliency precision decrease, while all-saliency precision increases, indicating a shift toward more concentrated explanations under stronger supervision.}
    \label{fig:WeightAccuracy_and_SaliencyPrecision}
\end{figure*}

\subsection{Effect of Explanation Loss Coefficient ($\alpha$)}

\subsubsection{Predictive Performance}
We first analyze the effect of the explanation loss coefficient on predictive performance. 
Fig.~\ref{fig:WeightAccuracyCoverage} summarizes the distribution of test accuracy across four explanation loss coefficient values using box plots, capturing median performance, variability, and mean accuracy.
As shown in Fig.~\ref{fig:WeightAccuracyCoverage}, the median test accuracy remains relatively stable across all explanation loss coefficients, indicating that explanation supervision does not systematically degrade predictive performance. 
However, the mean accuracy exhibits a mild downward trend as the explanation loss coefficient increases, suggesting that stronger explanation supervision can adversely affect a subset of runs.

In addition, the spread of test accuracy increases for larger coefficient values, reflecting greater variability in model performance across disease categories. 
This increased variance indicates that while many models retain strong predictive performance under explanation supervision, certain tasks are more sensitive to stronger explanation constraints. Overall, these results suggest that explanation supervision introduces a modest accuracy–variability trade-off rather than a uniform degradation in predictive performance.

\subsubsection{Explanation Metrics}
Fig.~\ref{fig:SaliencyPrecisionByWeights} presents the corresponding trends in annotation coverage, top-saliency precision, and all-saliency precision under different explanation loss coefficients.
Explanation-related metrics exhibit systematic changes as the explanation loss coefficient increases. Fig.~\ref{fig:SaliencyPrecisionByWeights} shows that annotation coverage is highest at the lowest coefficient ($\alpha = 0.25$) and decreases as the coefficient increases. A similar trend is observed for top-saliency precision, indicating that weaker explanation supervision produces broader saliency maps whose strongest activations are more likely to intersect coarse annotation boxes.

Conversely, all-saliency precision increases monotonically with the explanation loss coefficient. This trend indicates that stronger explanation supervision suppresses diffuse activations and redistributes saliency mass toward expert-annotated regions. As a result, although fewer annotated regions are highlighted at higher coefficients, a larger proportion of the overall saliency mass lies within clinically relevant areas.

Taken together, these results demonstrate that increasing the explanation loss coefficient shifts model behavior from broad spatial coverage toward more selective and concentrated explanations, while maintaining competitive classification accuracy.

% \subsection{Trade-off Between Annotation Coverage and Saliency Precision}

We next examine the relationship between annotation coverage and saliency concentrations. As shown in Fig.~\ref{fig:SaliencyPrecisionByWeights}, lower explanation loss coefficients yield higher annotation coverage and higher top-saliency precision. In this regime, Grad-CAM heatmaps are spatially diffuse, resulting in broad overlap with coarse bounding boxes and increasing the likelihood that the most salient pixels fall within annotated regions. As the explanation loss coefficient increases, annotation coverage and top-saliency precision decrease, indicating that saliency becomes more selective and activates fewer annotated regions. This behavior reveals a clear trade-off controlled by the explanation loss coefficient. Lower coefficients favor broad spatial coverage and diffuse explanations, whereas higher coefficients promote concentrated and spatially faithful saliency at the cost of reduced coverage. Importantly, this trade-off allows explanation behavior to be tuned according to annotation granularity and interpretability requirements, without substantially affecting classification accuracy.

\subsection{Effect of Explanation Loss Formulation}

\begin{figure*}[t]
    \centering
    \subfloat[Test accuracy across loss formulations]{
        \includegraphics[width=0.48\textwidth, height=5cm]{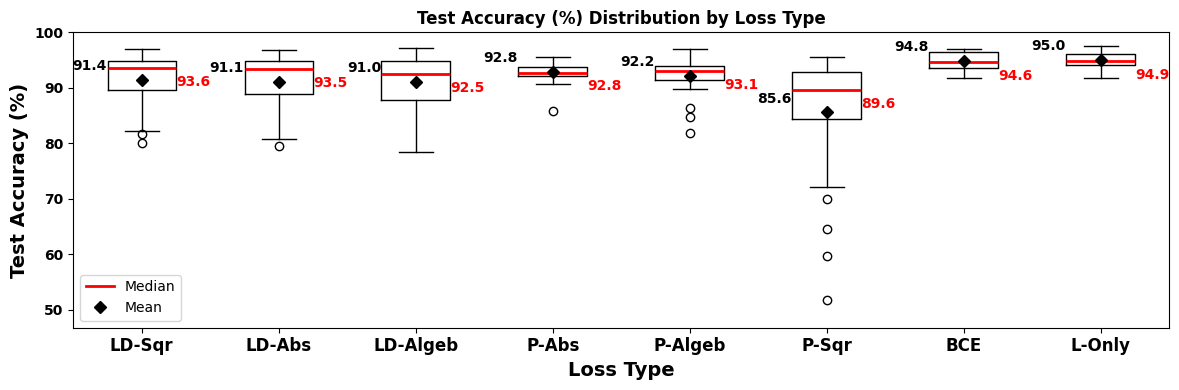}
        \label{fig:testAccuracyByLossType}
    }
    \hfill
    \subfloat[Annotation coverage behavior]{
        \includegraphics[width=0.48\textwidth, height=5cm]{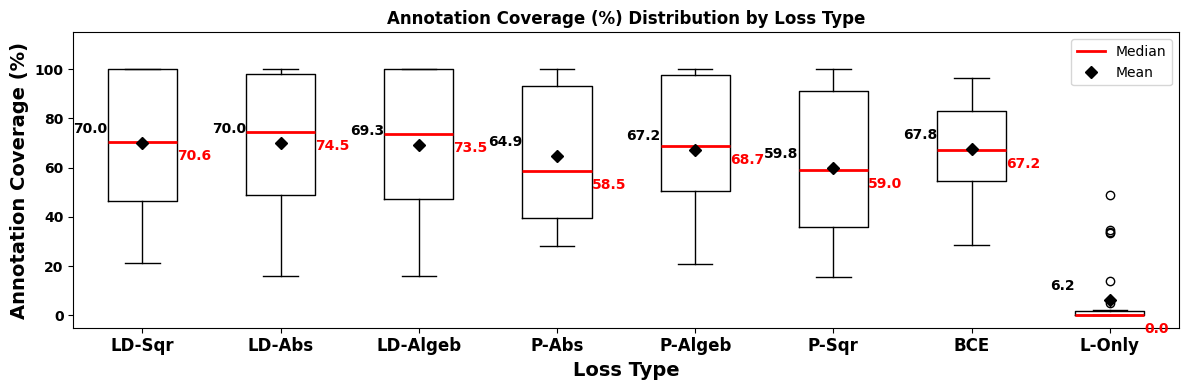}
        \label{fig:BoxCoverageByLossType}
    }\\[6pt]

    \subfloat[Top-saliency precision]{
        \includegraphics[width=0.48\textwidth, height=5cm]{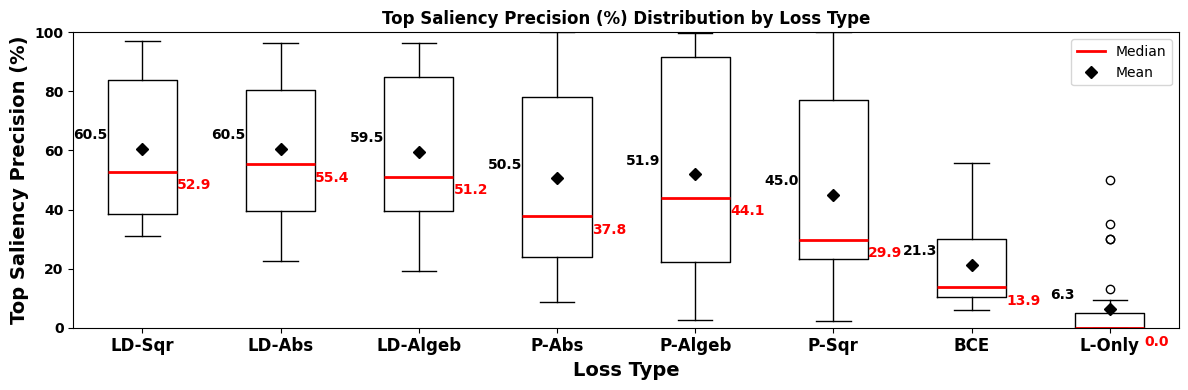}
        \label{fig:TopSaliencyByLossType}
    }
    \hfill
    \subfloat[All-saliency precision]{
        \includegraphics[width=0.48\textwidth, height=5cm]{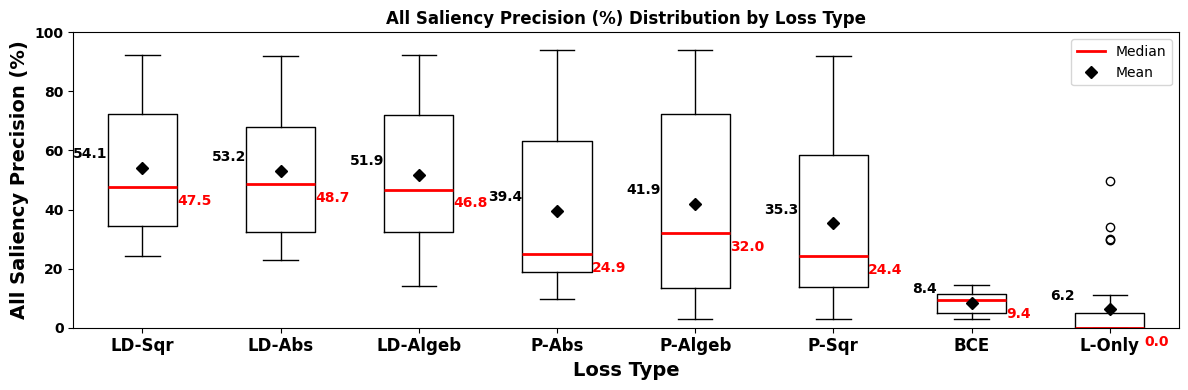}
        \label{fig:AllSaliencyPrecisionByLossType}
    }
    \caption{Effect of explanation loss formulation on predictive performance and explanation quality.
    Loss types are denoted using compact abbreviations on the X-axis.
    \textbf{LD} and \textbf{P} denote logit-difference and probability-difference explanation losses, respectively;
    \textbf{Abs}, \textbf{Algeb}, and \textbf{Sqr} indicate absolute, algebraic, and squared variants.
    \textbf{L-Only} applies explanation supervision directly to the target class logit without pairwise contrast.
    (a) Distribution of test accuracy across loss formulations, showing comparable accuracy across most explanation-aware objectives, with no substantial separation in median performance.
    (b) Distribution of annotation coverage, measuring how often saliency maps acknowledge expert-annotated regions under coarse bounding-box supervision.
    (c) Top-saliency precision, measuring the alignment of the most salient regions with expert-annotated areas.
    (d) All-saliency precision, capturing the overall concentration of attribution mass within annotated regions.
    Together, the results indicate that while accuracy remains largely insensitive to loss formulation, explanation quality is strongly affected, with logit-based squared losses yielding the most consistent and spatially faithful explanations.}
    \label{fig:LossTypeAblation}
\end{figure*}

We next analyze how different explanation loss formulations influence predictive performance and explanation quality. 
Fig.~\ref{fig:LossTypeAblation} summarizes the distributions of test accuracy and explanation metrics across loss types using box plots, enabling comparison of central tendency and variability across disease categories.

Across explanation loss formulations, the distributions of test accuracy in Fig. \ref{fig:testAccuracyByLossType} show no substantial separation between loss types, indicating that predictive performance is largely insensitive to the specific explanation loss design. 
With the exception of the standard BCE and \textit{Logit Only} baselines, which are examined primarily in terms of their explanation behavior, most explanation-aware objectives achieve comparable accuracy distributions. 
% Close inspection revels that the spread of accuracies for P-Abs and P-Algeb is smaller than others, most of the time the median accuracy values higher than their mean. This indicates that across different experimental settings, model performed relatively better.

Probability-based absolute and algebraic formulations exhibit slightly tighter interquartile ranges and marginally higher mean accuracy values compared to logit-based losses. 
However, these gains are not reflected at the median level, where logit-based formulations achieve comparable or slightly higher central performance. 
The squared probability-difference formulation performs poorly in terms of accuracy stability, exhibiting wider dispersion and lower central tendency. Overall, these results indicate that differences in explanation loss formulation do not materially affect classification accuracy, and that observed performance variations are minor relative to the pronounced differences in explanation quality discussed below.

Substantial differences emerge when examining explanation quality in Figs.~\ref{fig:BoxCoverageByLossType}, 
\ref{fig:TopSaliencyByLossType}, and 
\ref{fig:AllSaliencyPrecisionByLossType} across loss formulations. Models trained using standard BCE loss achieve relatively high annotation coverage due to spatially diffuse saliency maps that broadly activate across the image. 
While this behavior increases the likelihood of overlapping coarse annotation boxes, it results in poor saliency precision, as attribution mass is spread over clinically irrelevant regions.

The LogitOnly objective performs poorly across all explanation metrics. 
Because this formulation lacks contrastive information between positive and negative class evidence, the resulting saliency maps are weak and unstable, leading to both low coverage and low saliency precision. Among explanation-aware objectives, logit-based losses consistently outperform probability-based alternatives in saliency precision metrics. 
In particular, the squared logit-difference formulation exhibits the tightest distributions across annotation coverage, top-saliency precision, and all-saliency precision, indicating both strong localization and stable explanation behavior.

This behavior may be explained by the properties of logit-based supervision. Unlike probability outputs, which saturate near 0 and 1 and yield diminishing gradients, logits provide smoother and more informative gradients throughout training. 
Furthermore, the squared penalty amplifies deviations from the desired contrastive behavior, more strongly discouraging diffuse activations and encouraging concentrated attribution within annotated regions.
These findings highlight that explanation-aware training effectiveness is sensitive to loss formulation, and that logit-based, contrastive supervision provides a principled balance between predictive performance and spatial faithfulness of explanations.

\subsection{Summary of Quantitative Findings}

All reported performance and explanation metrics are summarized across seven disease categories, encompassing both disease-positive and disease-negative samples. 
Across all diseases, consistent trends are observed with respect to explanation loss formulation and supervision strength, $\alpha$, as reflected in the distributions of accuracy and explanation metrics.
In particular, logit-based explanation losses consistently yield higher saliency precision than probability-based alternatives, while squared penalties provide the strongest spatial alignment with annotated regions.

Similarly, the coverage--precision trade-off governed by the explanation loss coefficient is consistently observed across disease categories, indicating that the effect of explanation strength is not driven by any single pathology. 
Lower coefficients favor broader spatial coverage of coarse annotations, whereas higher coefficients promote more localized and focused explanations, with only modest changes in median predictive accuracy.
Taken together, these results demonstrate that the observed behaviors of explanation loss design and weighting are robust across diverse clinical findings and disease categories. 
This consistency across distributions supports the general applicability of the proposed explanation-aware training framework for medical image classification under noisy and heterogeneous annotation conditions.

\section{Conclusion}
This work introduces a unified explanation-aware training approach for medical image classification that jointly optimizes predictive performance and explanation quality by augmenting standard cross entropy loss with multiple explanation-driven objectives. 
By design, the proposed framework enables systematic analysis of how explanation supervision influences both model predictions and saliency behavior within a single end-to-end training pipeline.
Through extensive quantitative evaluation, we provide a principled characterization of the trade-off between explanation loss formulation, supervision strength, and explanation quality. 
Across all evaluated objectives, predictive performance remains comparable at the distributional level; however, clear differences emerge in explanation behavior. 
% In particular, the squared logit-difference formulation consistently achieves the most favorable and stable performance across annotation coverage, top-saliency precision, and all-saliency precision metrics, indicating superior spatial alignment and robustness of explanations.
We further show that the explanation loss coefficient serves as an effective control parameter for balancing accuracy and interpretability. 
% Lower supervision strengths favor stable predictive performance, higher annotation coverage, and improved top-saliency precision, while increasing the coefficient progressively encourages more localized and concentrated explanations. 

These trends reveal a tunable coverage– saliency concentration trade-off that can be adjusted according to clinical requirements without substantially altering median classification accuracy.
% Finally, we demonstrate that the observed effects of explanation loss design and weighting are consistent across seven disease categories, establishing the robustness of the proposed framework under heterogeneous and noisy annotation conditions. 
Findings validate explanation-aware training as a practical and generalizable strategy for improving the faithfulness of visual explanations in medical imaging, while offering concrete guidance for selecting loss formulations and supervision strengths in real-world clinical settings. In the future we plan to extend this work by incorporating multi-class and multi-label classification problems as well as investigating the proposed approach on other types of biomedical imaging problems. 

\section*{References}

\bibliographystyle{IEEEtran}
\bibliography{references}

@article{loh2022xaihealthcare,
  title={Application of explainable artificial intelligence for healthcare: A systematic review of the last decade (2011--2022)},
  author={Loh, Hwee W. and Ooi, Chee P. and Seoni, S. and Barua, P. D. and Molinari, F. and Acharya, U. R.},
  journal={Computer Methods and Programs in Biomedicine},
  volume={226},
  pages={107161},
  year={2022}
}

@inproceedings{selvaraju2017gradcam,
  title={Grad-CAM: Visual explanations from deep networks via gradient-based localization},
  author={Selvaraju, Ramprasaath R. and others},
  booktitle={Proceedings of the IEEE International Conference on Computer Vision},
  year={2017}
}

@inproceedings{adebayo2018sanity,
  title={Sanity checks for saliency maps},
  author={Adebayo, Julius and others},
  booktitle={Advances in Neural Information Processing Systems},
  year={2018}
}

@incollection{kindermans2017reliability,
  title={The (Un)reliability of Saliency Methods},
  author={Kindermans, Pieter-Jan and Hooker, Sara and Adebayo, Julius and Alber, Maximilian and Sch{\"u}tt, Kristof T. and D{\"a}hne, Sven and Erhan, Dumitru and Kim, Been},
  booktitle={Explainable AI: Interpreting, Explaining and Visualizing Deep Learning},
  pages={267--280},
  publisher={Springer},
  year={2019},
  doi={10.1007/978-3-030-28954-6_14}
}

@inproceedings{ross2017right,
  title={Right for the right reasons: Training differentiable models by constraining their explanations},
  author={Ross, Andrew S. and Hughes, Michael C. and Doshi-Velez, Finale},
  booktitle={Proceedings of IJCAI},
  year={2017}
}

@inproceedings{sun2020explanation,
  title={Explanation-guided training for cross-domain few-shot classification},
  author={Sun, J. and others},
  booktitle={Proceedings of ICPR},
  year={2020}
}

@article{uddin2025expert,
  title={Expert-guided explainable few-shot learning for medical image diagnosis},
  author={Uddin, I. I. and Wang, L. and Santosh, K.},
  journal={arXiv preprint arXiv:2509.08007},
  year={2025}
}

@article{saporta2022evaluation,
  title={Benchmarking saliency methods for chest X-ray interpretation},
  author={Saporta, A. and others},
  journal={Nature Machine Intelligence},
  year={2022}
}

@inproceedings{li2018thoracic,
  title={Thoracic disease identification and localization with limited supervision},
  author={Li, Z. and others},
  booktitle={Proceedings of CVPR},
  year={2018}
}

@inproceedings{ribeiro2016lime,
  title     = {``Why Should I Trust You?'': Explaining the Predictions of Any Classifier},
  author    = {Ribeiro, Marco Tulio and Singh, Sameer and Guestrin, Carlos},
  booktitle = {Proceedings of the 22nd ACM SIGKDD International Conference on Knowledge Discovery and Data Mining},
  pages     = {1135--1144},
  year      = {2016},
  publisher = {ACM}
}

@inproceedings{lundberg2017shap,
  title     = {A Unified Approach to Interpreting Model Predictions},
  author    = {Lundberg, Scott M. and Lee, Su-In},
  booktitle = {Advances in Neural Information Processing Systems (NeurIPS)},
  volume    = {30},
  year      = {2017}
}

@article{lapuschkin2019lrp,
  title   = {Unmasking Clever Hans Predictors and Assessing What Machines Really Learn},
  author  = {Lapuschkin, Sebastian and W{\"a}ldchen, Stephan and Binder, Alexander and Montavon, Gr{\'e}goire and Samek, Wojciech and M{\"u}ller, Klaus-Robert},
  journal = {Pattern Recognition},
  volume  = {87},
  pages   = {1--15},
  year    = {2019}
}

@article{sefcik2021lrp,
  title   = {Explaining decisions of convolutional neural networks for brain tumor classification},
  author  = {\v{S}ef\v{c}\'{i}k, Martin and Samek, Wojciech and Lapuschkin, Sebastian and Binder, Alexander and M{\"u}ller, Klaus-Robert},
  journal = {Pattern Recognition},
  volume  = {112},
  pages   = {107797},
  year    = {2021}
}

@article{caragliano2021doctor,
  title   = {Doctor-in-the-Loop: Human-in-the-Loop Framework for Clinical Decision Support Using Deep Learning},
  author  = {Caragliano, Antonio and Bhandari, Arjun and Lakshminarayanan, Vasudevan and others},
  journal = {IEEE Transactions on Medical Imaging},
  volume  = {40},
  number  = {5},
  pages   = {1328--1340},
  year    = {2021}
}

@article{panwar2020deep,
  title={A deep learning and Grad-CAM based color visualization approach for fast detection of COVID-19 cases using chest X-ray and CT-scan images},
  author={Panwar, Harsh and Gupta, P. K. and Siddiqui, M. K. and Morales-Menendez, R. and Bhardwaj, P.},
  journal={International Journal of Imaging Systems and Technology},
  year={2020},
  doi={10.1002/ima.22427}
}

@article{umair2021detection,
  title={Detection of COVID-19 Using Transfer Learning and Grad-CAM Visualizations on Chest X-Ray Images},
  author={Umair, M. and others},
  journal={Sensors},
  volume={21},
  number={17},
  year={2021},
  doi={10.3390/s21175813}
}

@article{erukude2025explainable,
  title={Explainable deep learning in medical imaging: brain tumor and pneumonia detection},
  author={Erukude, Sai Teja and Marella, Viswa Chaitanya and Veluru, Suhasnadh Reddy},
  journal={arXiv preprint arXiv:2510.21823},
  year={2025}
}

@article{saporta2022benchmarking,
  title={Benchmarking saliency methods for chest X-ray interpretation},
  author={Saporta, A. and others},
  journal={Nature Machine Intelligence},
  volume={4},
  pages={269--281},
  year={2022}
}

@article{rajpurkar2017chexnet,
  title={CheXNet: Radiologist-Level Pneumonia Detection on Chest X-Rays with Deep Learning},
  author={Rajpurkar, Pranav and Irvin, Jeremy and Zhu, Kaylie and Yang, Brandon and Mehta, Hershel and Duan, Tony and Ding, Daisy and Bagul, Aarti and Langlotz, Curtis and Shpanskaya, Katie and others},
  journal={arXiv preprint arXiv:1711.05225},
  year={2017}
}

@article{litjens2017survey,
  title={A survey on deep learning in medical image analysis},
  author={Litjens, Geert and Kooi, Thijs and Bejnordi, Babak Ehteshami and Setio, Arnaud Arindra Adiyoso and Ciompi, Francesco and Ghafoorian, Mohsen and van der Laak, Jeroen AWM and van Ginneken, Bram and S{\'a}nchez, Clara I},
  journal={Medical Image Analysis},
  volume={42},
  pages={60--88},
  year={2017},
  publisher={Elsevier}
}

@article{zech2018variable,
  title={Variable generalization performance of a deep learning model to detect pneumonia in chest radiographs},
  author={Zech, John R and Badgeley, Marcus A and Liu, Manway and Costa, Anthony B and Titano, Joseph J and Oermann, Eric K},
  journal={PLOS Medicine},
  volume={15},
  number={11},
  pages={e1002683},
  year={2018},
  publisher={Public Library of Science}
}

@article{badgeley2019deep,
  title={Deep learning predicts hip fracture using confounding patient and healthcare variables},
  author={Badgeley, Marcus A and Zech, John R and Oakden-Rayner, Luke and Glicksberg, Benjamin S and Liu, Manway and Gale, William and McConnell, Michael V and Percha, Bethany and Snyder, Michael and Dudley, Joel T},
  journal={NPJ Digital Medicine},
  volume={2},
  number={1},
  pages={31},
  year={2019},
  publisher={Nature Publishing Group}
}

@article{kelly2019key,
  title={Key challenges for delivering clinical impact with artificial intelligence},
  author={Kelly, Christopher J and Karthikesalingam, Alan and Suleyman, Mustafa and Corrado, Greg and King, Dominic},
  journal={BMC Medicine},
  volume={17},
  number={1},
  pages={195},
  year={2019},
  publisher={BioMed Central}
}

@article{tonekaboni2019clinicians,
  title={What clinicians want: contextualizing explainable machine learning for clinical end use},
  author={Tonekaboni, Sana and Joshi, Shalmali and McCradden, Melissa D and Goldenberg, Anna},
  journal={NPJ Digital Medicine},
  volume={2},
  number={1},
  pages={79},
  year={2019},
  publisher={Nature Publishing Group}
}

@misc{rieger2020interpretationsusefulpenalizingexplanations,
      title={Interpretations are useful: penalizing explanations to align neural networks with prior knowledge}, 
      author={Laura Rieger and Chandan Singh and W. James Murdoch and Bin Yu},
      year={2020},
      eprint={1909.13584},
      archivePrefix={arXiv},
      primaryClass={cs.LG},
      url={https://arxiv.org/abs/1909.13584}, 
}

@article{Nguyen2022VinDrCXR,
  title     = {VinDr-CXR: An open dataset of chest X-rays with radiologist annotations},
  author    = {Nguyen, Hoang T. and Pham, Hieu T. and Le, Trong T. and Nguyen, Hieu Q. and Vu, Minh T. and others},
  journal   = {Scientific Data},
  volume    = {9},
  number    = {1},
  pages     = {429},
  year      = {2022},
  publisher = {Nature Publishing Group},
  doi       = {10.1038/s41597-022-01498-6}
}

@inproceedings{Huang2017DenseNet,
  title     = {Densely Connected Convolutional Networks},
  author    = {Huang, Gao and Liu, Zhuang and Van Der Maaten, Laurens and Weinberger, Kilian Q.},
  booktitle = {Proceedings of the IEEE Conference on Computer Vision and Pattern Recognition (CVPR)},
  pages     = {4700--4708},
  year      = {2017}
}

@article{rajaraman2024transfercxr,
  title={Impact of transfer learning strategies on the generalization of deep learning models for chest radiograph classification},
  author={Rajaraman, Sivaramakrishnan and Antani, Sameer and Candemir, Sema and Xue, Zhiyun and Thoma, George R.},
  journal={PLOS Digital Health},
  volume={3},
  number={1},
  pages={e0000418},
  year={2024},
  publisher={Public Library of Science},
  doi={10.1371/journal.pdig.0000418}
}

@misc{faruqui2025explainabilitycnnbasedclassification,
      title={Explainability of CNN Based Classification Models for Acoustic Signal}, 
      author={Zubair Faruqui and Mackenzie S. McIntire and Rahul Dubey and Jay McEntee},
      year={2025},
      eprint={2509.08717},
      archivePrefix={arXiv},
      primaryClass={cs.SD},
      url={https://arxiv.org/abs/2509.08717}, 
}

@article{doshi2017rigorous,
  title   = {Towards a Rigorous Science of Interpretable Machine Learning},
  author  = {Doshi-Velez, Finale and Kim, Been},
  journal = {arXiv preprint arXiv:1702.08608},
  year    = {2017}
}

@article{guidotti2018survey,
  title   = {A Survey of Methods for Explaining Black Box Models},
  author  = {Guidotti, Riccardo and Monreale, Anna and Ruggieri, Salvatore and Turini, Franco and Pedreschi, Dino and Giannotti, Fosca},
  journal = {IEEE Access},
  volume  = {6},
  pages   = {61538--61565},
  year    = {2018}
}

@article{samek2019xai,
  title   = {Explainable Artificial Intelligence: Understanding, Visualizing and Interpreting Deep Learning Models},
  author  = {Samek, Wojciech and Wiegand, Thomas and M{\"u}ller, Klaus-Robert},
  journal = {ITU Journal: ICT Discoveries},
  volume  = {1},
  number  = {1},
  year    = {2019}
}

@article{tjoa2020medical,
  title   = {A Survey on Explainable Artificial Intelligence (XAI): Toward Medical XAI},
  author  = {Tjoa, Erico and Guan, Cuntai},
  journal = {IEEE Transactions on Neural Networks and Learning Systems},
  volume  = {32},
  number  = {11},
  pages   = {4793--4813},
  year    = {2021}
}

\end{document}